# MARC: A multi-agent robots control framework for enhancing reinforcement learning in construction tasks

Kangkang Duan; Christine Wun Ki Suen; Zhengbo Zou *

Department of civil engineering, The University of British Columbia, Vancouver, BC, Canada.

*Corresponding author, email: zhengbo@civil.ubc.ca

**Abstract:**
Letting robots emulate human behavior has always posed a challenge, particularly in scenarios involving multiple robots. In this paper, we presented a framework aimed at achieving multi-agent reinforcement learning for robot control in construction tasks. The construction industry often necessitates complex interactions and coordination among multiple robots, demanding a solution that enables effective collaboration and efficient task execution. Our proposed framework leverages the principles of proximal policy optimization and developed a multi-agent version to enable the robots to acquire sophisticated control policies. We evaluated the effectiveness of our framework by learning four different collaborative tasks in the construction environments. The results demonstrated the capability of our approach in enabling multiple robots to learn and adapt their behaviors in complex construction tasks while effectively preventing collisions. Results also revealed the potential of combining and exploring the advantages of reinforcement learning algorithms and inverse kinematics. The findings from this research contributed to the advancement of multi-agent reinforcement learning in the domain of construction robotics. By enabling robots to behave like human counterparts and collaborate effectively, we pave the way for more efficient, flexible, and intelligent construction processes.

**1. Introduction**
In the construction industry, although there are still limited applications, robots provide the paradigm for future construction activities [1-3]. Just like buildings evolve to be more resilient and comfortable with new materials and design theorem, construction tends to become convenient and efficient with emerging artificial intelligence techniques. Current construction tasks such as structural assembly, earth-moving, and hoist control are labor-intensive, time-consuming, and error-prone activities that not only limit efficiency due to the weakness of the human body but also result in thousands of accidents every year [4-5].

Towards more intelligent and reliable mechanical devices, studies have investigated and developed diverse mechanical and machine learning theories so that these electric-driven steel creatures can finally replace human workers [4-5]. Early research work[8-10] has demonstrated the potential of using robots to complete fundamental construction tasks such as structural assembly. New robots [11-13] are developed based on the specific requirements of construction activities. For example, Vega-Heredia et al. [13] designed a cleaning robot that can move on the windows. Saltaren et al. [12] developed a parallel robot that can climb metal structures. Another urgent problem exists in the control of the robot. Although experts can remotely manipulate a robot [11], the perfect solution should be autonomous construction robots that can appropriately respond to the environment.

Proportion integration differentiation (PID) [8], inverse kinematics [12], and path planning [14] have been developed to code the control system of construction robots. In recent years, the control method named reinforcement learning (RL) based on the Markov decision process (MDP) has drawn enormous interest from the research community. RL shows a promising generalization ability in comparison with other methods [15]. In RL, we assume the robot perceives the state, adopts an action based on the state, and receives an immediate reward signal for each step [16]. The goal of RL is to learn a policy that can maximize the accumulated rewards in an episode. The policy maps the states onto the actions, so for large action and state space the policy will be complex. Researchers applied neural networks to represent the policy to be optimized and developed the deep Q-learning (DQN)[17] where a neural network was used to approximate the Q value function $Q(s, a)$. $Q(s, a)$ is the value that the robot can earn at state $s$ choosing action $a$. After that, policy gradient algorithms were proposed for continuous action and state space. Policy gradient algorithms directly utilized a deep neural network to represent the policy and used the experience during exploring the environment to update the network [16]. Algorithms such as proximal policy optimization (PPO)[18] and deep deterministic policy gradient (DDPG) [19] are typical policy gradient algorithms.

These RL methods are expected to succeed in control tasks of the majority of construction robots since many construction devices such as cranes have simple kinematics models and work in low dimensional space. Related studies also demonstrate the feasibility of employing RL in construction tasks such as structural assembly [20-21], hoist control [15], and material transportation [22]. In these studies, RL shows its remarkable generalization ability and learning performance in comparison with human control or other conventional control methods.

However, the research is still limited: rare applications in realistic construction sites; limitations in the fusion of multiple sensors; simple learning environments. For example, all the applications focus on a single-agent learning environment. However, realistic construction tasks involve collaboration and competition between robots in the multi-agent environment. Multi-agent environments bring challenges to learning since the environment is non-stationary from the perspective of individual agents, which casts a shadow over the convergence of learning [23-24]. Another key question in a multi-agent construction environment is how to work together to complete a task without collisions. Researchers have proposed various algorithms and techniques to address these challenges in MARL, such as multi-agent extensions of DQN [25], policy gradient methods (e.g., MADDPG - Multi-Agent DDPG [24]), and value decomposition methods (e.g., QMIX - Q-learning with a Mixer [26]). In this paper, we proposed a framework for multi-agent reinforcement learning (MARL) in construction robot control and further applied the framework to four collaborative construction tasks to evaluate the learning performance, the main contributions are as follows:
1. We have developed a cutting-edge MARL framework for the control of construction robots. Our framework incorporates a streamlined version of Multi-agent PPO, which has demonstrated exceptional learning capabilities and performance.
2. We have designed a range of learning environments specifically tailored for construction robots within our framework. These environments encompass the essential components of a construction setting, including robot models, various instruments, and realistic construction scenes. Furthermore, our framework allows for seamless expansion to accommodate diverse construction scenarios beyond the standard model.
3. Our research has showcased the tremendous potential of merging RL with inverse kinematics, resulting in a promising synergy. By leveraging the speed and precision of inverse kinematics for joint action planning, coupled with the superior generalization capabilities of RL policies, we have demonstrated a hybrid approach that optimizes robot movement.
4. We conducted experimental tests across four construction tasks. The results highlighted the exceptional learning performance achieved when controlling multiple agents with a high degree of freedom, enabling seamless collaboration in task completion.

In the next section, we first introduced the related background; then, we explained our methodologies, which include the computer vision part, the basic information of MDP and RL, and the MARL part. After that, we built the task environments and gave detailed information on the learning configurations and tasks. Finally, we discussed the learning results and the effectiveness of our framework.

## 2. Related work

### 2.1 Reinforcement learning in construction robots

Robot-aided construction has been studied as the future paradigm of automatic construction for many years [8,12,14,27-28]. However, previous control methods based on PID [8], inverse kinematics [12,27], and path planning [14,28] suffer from low generalization ability in environment changes and weakness in solving collision problems. Emerging RL algorithms such as DQN [17], DDPG [19], PPO [18], and actor-critic [29] provide promising solutions to the aforementioned problems. Although the studies are currently limited, recent studies in construction robots have demonstrated the feasibility of RL-based control [3].

RL can either be used to control current construction devices such as hoists and vehicles or be used to control robots to help humans complete tasks or replace workers. For example, Santos et al. [22] adopted finite action-set learning automata to control a quad-rotor robot for component transportation in a beam assembly task. Lee et al. [15] demonstrated that using RL methods such as DQN can significantly enhance the efficiency of construction hoist control. Azad et al. [30] applied DQN to solve challenges brought about by the changes in robot dynamics in the existence of heavy external loads and disturbances in performing construction tasks. Belousov et al. [20] and Apolinarska et al. [21] utilized the DDPG algorithm to train a 6-degree-of-freedom robot arm to do structural assembly tasks. In addition, Liang and Lei et al. [31-32] manifested the improvement in learning efficiency by using expert demonstrations in RL to train a robot arm to finish construction tasks.

However, previous applications are essentially learning a construction task in single-agent environments. Many construction tasks such as structural assembly actually involve the collaboration or competition between multiple agents. For a multi-agent environment, one agent needs to consider the influence of other agents, which brings new problems such as avoiding collisions between agents. In this study, we developed a concise version of the multi-agent PPO algorithm and further evaluated the learning performance in multi-agent construction tasks.

### 2.2 Multi-agent reinforcement learning in construction and building management

Despite many construction activities such as beam installation and vehicle scheduling involve the collaboration or competition of multiple agents, studies on this topic [33] are still limited. Previous studies [33-36] mainly focus on decision-making in infrastructure and building management. Since the management of buildings and infrastructure is a complex issue involving different components (e.g., different parts of a bridge [34]), MARL has been applied to solve the curse of dimensionality brought by multiple agents. For example, Andriotis et al. [34] proposed the deep centralized multi-agent actor-critic method and adapted it to make decisions such as inspection, minor repair, and major repair in bridge maintenance. In this work, the states and actions of each component were discretized into several values. Asghari et al. [33] applied DQN and advantage actor-critic to the management of infrastructure. They first built a microworld considering the uncertainties and managerial considerations for simulating the response of infrastructure management, then followed by the MARL algorithms for decision making. Jose et al. [36] utilized MADDPG to control the energy consumption in ten buildings to minimize the total cost of cooling. The action is the energy buildings store or release in every hourly interval and the state includes the hour of the day, the outdoor temperature, and the state of charge in every chilled water tank. Results showed that MADDPG control outperformed the rule-based controller. Yu et al. [35] proposed the attention-based multi-agent deep reinforcement learning algorithm for the energy consumption minimization problem of buildings. In this work, the attention mechanism was employed to quantify the contributions from other agents when training one agent's value neural networks.

In addition, the research work of Miyazaki et al. [37] demonstrated the feasibility of using MADDPG in vehicle scheduling at construction sites. Liu et al. [38] treated the rebar design in reinforced concrete structures as a path-planning problem where each rebar was considered as an agent and applied multi-agent DQN to generate the clash-free design.

However, gaps also exist in previous MARL applications. For example, previous studies tended to discretize the state and action space or use low dimensional action space, which can simplify the task but also limit the application for the general control of continuous action/state systems such as robots. Besides, some studies only simply employed RL algorithms such as DQN on multiple agents without considering the communication between agents. According to the research [24], the non-stationary environment from the perspective of individual agents limits the performance of learning. Construction robot control raises a challenge in this topic due to the high dimensional and continuous state and action spaces. Although many construction tasks require collaboration between robots and applying more agents can improve efficiency, using MARL methods to control robots to finish construction tasks is still a whiteboard.

## 3. Methods

As shown in Fig. 1, the agents interact with the environment and first receive the state information from their observation. Then, agents can either select MARL or inverse kinematics to generate actions based on the state information: agents apply MARL to move to target positions and then utilize inverse kinematics to finally complete the task accurately.

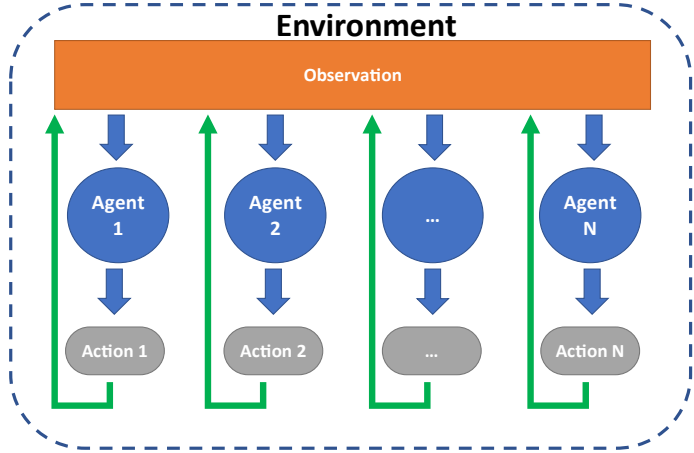

Fig. 1 Overview of our multi-agent reinforcement learning algorithm.

The combination of MARL and inverse kinematics first leverages the generalization ability and collision avoidance ability of RL. To prevent collisions of links of robots is a hard problem in inverse kinematics, especially in such a dynamic environment including multiple agents. On the other hand, as some algorithms such as PPO utilize stochastic policy (i.e., for a given state, the output actions are sampled from a distribution), using inverse kinematics can accurately pick up and place objects to target positions.

The inverse kinematics part uses the IKFast tool proposed by Diankov[39]. IKFast generates analytical solutions for inverse kinematics problems of manipulators. It uses a symbolic representation of the robot's kinematics and derives closed-form solutions.

### 3.1 Reinforcement learning background

In this paper, we consider continuous space MDPs that are represented by the tuple: $(\mathcal{S}, \mathcal{A}, \mathcal{P}, r, \gamma, s_0)$, where $\mathcal{S}$ is a set of continuous spaces; $\mathcal{A}$ is a set of continuous actions; $\mathcal{P}$ is the transition probability; $r$ is the reward; $\gamma$ is the discount factor; $s_0$ is the initial state. Common reinforcement learning algorithms such as policy gradient attempt to learn a policy $\pi: \mathcal{S} \times \mathcal{A} \to \mathbb{R}$ that can maximize the accumulated rewards $\sum_{t=0}^{T-1} \gamma^t r_t(s_t, a_t)$. The reward $r_t$ at timestep, $t$ can be obtained by executing the action $a_t$ at state $s_t$ in the environment.

Policy gradient algorithms apply a neural network to represent the policy $\pi_\theta$, where $\theta$ is the parameter of the network. To maximize the reward accumulation $J(\theta)$ under policy $\pi_\theta$ in one episode, policy gradient aims to adopt the gradient decent methods to optimize $\theta$ [16]:

$$\nabla J(\theta) \propto \mathbb{E}_\pi Q_\pi(s, a) \nabla_\theta \ln \pi(a|s, \theta) \quad (1)$$

Where $Q_\pi(s, a)$ is the value one agent can receive at state $s$ by executing action $a$ under policy $\pi$; $\pi(a|s, \theta)$ is the probability of executing action $a$ under state $s$ by following policy $\pi_\theta$. Some policy gradient algorithms such as PPO improve the learning performance by replacing the value function $Q_\pi$ with an advantage function $\hat{A}$ and restricting the changes from the old policy to the new policy (see [18]).

### 3.2 Multi-agent reinforcement learning control

When there is one agent, the environment is considered stationary without the influence of other agents: executing action $a$ at state $s$ will always lead to a fixed state $s'$. In this study, there are at least two agents: the right-arm agent and the left-arm agent. These two agents need to collaborate or compete while avoiding collision with each other. Such an environment is called a multi-agent environment in RL.

The key factor to scaling common RL algorithms to environments with multiple agents is the observations of agents. Traditional single-agent RL methods such as DQN and policy gradient perform unpromisingly in multi-agent environments due to the non-stationary environment from the perspective of any individual agent (if they only have partial information of the environment). For example, if we simply apply DQN to each agent $i$ in an environment, the reliability of experience tuples $(s, a, r, s')$ in the reply buffer in DQN faces challenges because $P(s'|s, a, \pi_1, ..., \pi_N) \neq P(s'|s, a, \pi_1', ..., \pi_N')$ when the policy $\pi_i$ of agent $i$ changes to $\pi_i'$. $s$ is the state, $a$ is the action, $r$ denotes the reward, and $s'$ is the state of the next step. For policy gradient methods such as PPO, the challenge is the high variance brought by partial observations. According to the work of Lowe et al [24], a solution to multiple agents learning is sharing the actions taken by all agents, which makes the environment stationary even as the policies change.

Therefore, in this work, we scale PPO to multi-agent environments by sharing actions of all agents, which means, compared with traditional PPO [18], the "surrogate" objective function in this study can be written as:

$$L_{clip}(\theta_i) = \widehat{\mathbb{E}}_t \left[ \min\left( \frac{\pi_{\theta_i}(a_i|s)}{\pi_{\theta_{i,old}}(a_i|s)} \hat{A}_{i,t}, clip\left( \frac{\pi_{\theta_i}(a_i|s)}{\pi_{\theta_{i,old}}(a_i|s)}, 1 - \zeta, 1 + \zeta \right) \hat{A}_{i,t} \right) \right] \quad (2)$$

Where $t$ is the timestep; $\theta_i$ is the parameter for policy neural networks $\pi_{\theta_i}$; $\pi_{\theta_i}$ is the policy of agent $i$; $\pi_{\theta_{i,old}}$ is the policy of agent $i$ before the update; $\pi_{\theta_i}(a_i|s)$ is the probability of taking action $a_i$ under state $s$ for agent $i$; $\hat{A}_{i,t}$ denotes the advantages of agent $i$ at timestep $t$; $\zeta$ represents the clip coefficient aims to limit the value of $\frac{\pi_{\theta_i}(a_i|s)}{\pi_{\theta_{i,old}}(a_i|s)}$ between $1 - \zeta$ and $1 + \zeta$. In Eq. (2), agents share their actions via state $s$. As a result, we make makes the environment stationary when we compute $\pi_{\theta_i}(a_i|s)$ and $\hat{A}_{i,t}$.

For each agent, we want to maximize Eq. (2) to maximize the accumulated rewards that agent $i$ obtains during interacting with the environment. First, we need to initialize two neural networks for each agent: the policy neural network $\theta_i$ and the value function neural network $\mu_i$. The value function is employed to predict the value of a state, which can be used to calculate the advantage $\hat{A}_{i,t}$ when an episode is not completed. Then, agents interact with the environment and related information including action $a_i$, state $s$, rewards $r_i$, values $v_i$, $\pi_{\theta_i}(a_i|s)$, and masks $m_i$ is added to the buffer. Masks are used to indicate if an episode is ended [40]. After a certain number of steps, the data in the buffer is randomly sampled for updating the policy and value function. Detailed steps for this process are shown in Table 1.

Table 1 Multi-agent proximal policy optimization algorithm

| **Algorithm: Multi-Agent Proximal Policy Optimization** |
| --- |
| Initialize actor networks with random weights $\theta_1$, $\theta_2$, and critic networks with random weights $\mu_1$, $\mu_2$<br>Hyperparameters: number of iterations *N*, buffer size *M*, number of mini-batches *n*, number of epochs *K*, learning rate *lr*, clipping parameter $\zeta$, weight coefficients $\lambda_1$, $\lambda_2$.<br>Reset the environment.<br>**for** *i* = 1,2, .. , *N* **do**:<br>   **for** *j* = 1,2, … , *M* **do**:<br>       Collect observations, rewards of agents, actions of agents, predicted values of states, masks, and distribution of action probability for the buffer by running the current policies of agents |

in the environment
   Compute the value of the next step, the value of each step $v'_i$, advantages $\hat{A}_{i,t}$ using generalized advantage estimation according to the rewards for each agent
   **for** each agent **do**:
      **for** $k = 1,2, \ldots , K$ **do**:
         Sample $M//n$ samples in the buffer as the mini-batch
         Compute the loss $L_{clip}$
         Compute the loss of value function $L_v$:
         $$L_v = \frac{1}{2}(v_i - v'_i)^2$$
         Compute the entropy loss $L_e$:
         $$L_e = \sum -\pi_{\theta_i}(a_i|s) \log \pi_{\theta_i}(a_i|s)$$
         Update the network parameters $\theta_1, \theta_2, \mu_1, \mu_2$ by optimizing $L_{clip} + \lambda_1 L_v + \lambda_2 L_e$ with $lr$
**end for**
Return the learned policy models with weights $\theta_1, \theta_2$

As implied by Eq. (2), the performance of learning depends on the calculation of advantages $\hat{A}$. A common approach in RL for computing the advantages of taking one action is called generalized advantage estimation (GAE) [41]. The GAE algorithm combines the Monte Carlo and temporal difference approaches to estimating the advantage function, which provides a balance between bias and variance in the estimation. The equation can be written as:

$$\hat{A}_t = \delta_t + (\gamma \xi) \delta_{t+1} + \cdots + (\gamma \xi)^{T-t-1} \delta_{T-1} \quad (3)$$
$$\delta_t = r_t + \gamma V(s_{t+1}) - V(s_t) \quad (4)$$

When $\xi = 1$, Eq. (3) reduces to a general equation for computing advantage. In addition, the value of each step $v'_i$ is computed based on the general equation, which can be written as:

$$v'_{i,t} = r_t + \gamma r_{t+1} + \cdots + \gamma^{T-t-1} r_{T-1} + \gamma^{T-t} V(s_T) \quad (5)$$

It is noted that, during calculating Eq. (3) ~ (5), the mask $m_i$ is applied to indicate if the results of this step will be affected by the results of the previous step.

**3.3 Construction environment development**

In this study, we develop the basic construction environment based on the pybullet physical simulator[42]. To develop a construction environment in the pybullet physical simulator[42], the unified robotic description format (URDF) files are required. The simulator only provides very limited models so we need to develop lots of extra URDF objects to generate a construction scenario.

In this study, we utilize the model developed via 3D modeling software such as 3ds Max, Blender, and Cinema 4D. These models can be imported into Unity or other related software and then, we turn models into Object files with appropriate textures. After that, we use the Objective files as mesh files and build corresponding URDF files. In this simulator, all the objects should be imported as a URDF model. For some objects such as hammers, we basically define the objects only with a base link in the URDF file. As shown in Fig. 2, our framework includes the fundamental robot family and construction tool family developed based on the simulator models and other 3D models. The robot family includes humanoid robots, robot rams, and unmanned ground vehicles. Construction tools include fundamental instruments, beams, tables, and crates. New objects such as new buildings or construction site models can be added after being transferred into URDF files.

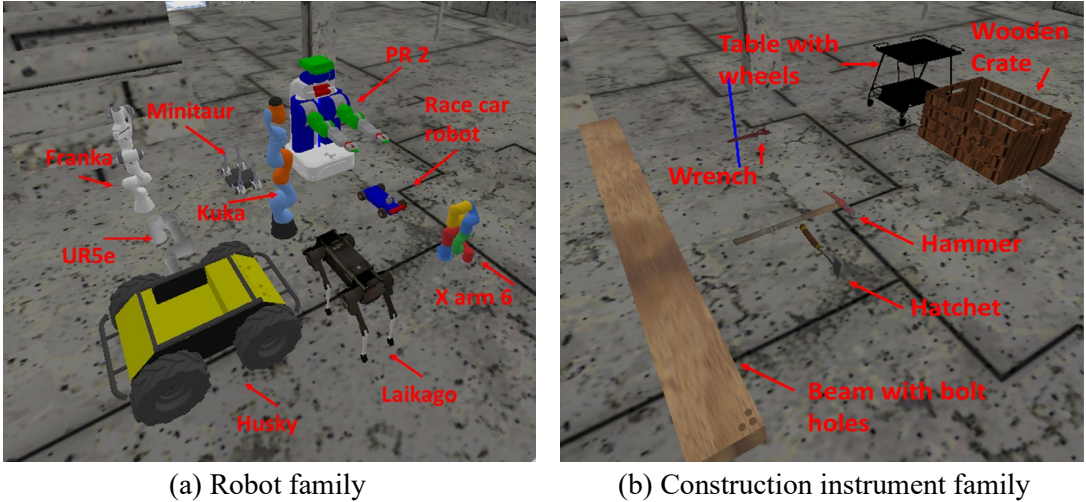

| (a) Robot family | (b) Construction instrument family |

Fig. 2 Elements for new environment development.

With these URDF models, construction scenarios can be established. We design the preliminary learning environments after the scenario is built. The environment includes the reward function, observation collection part, simulation execution part, and other necessary content such as environment reset and collision detection. To modify the environment towards a new learning task, one just needs to design the reward function and observation information.

## 4. Experimental tests

To test the performance of the proposed method, we developed multi-agent construction environments in the pybullet physical simulator[42]. Different collaboration and competition construction tasks can be further developed based on our basic environment.

### 4.1 Tasks statement

*Task 1* Multi-agent pick and place task (Fig. 3 a): The robot is tasked to pick up and place two bolts (denoted by cylinders: 0.1 m height, 0.008 m radius) on a platform (dimensions: 0.01 m×0.3 m×0.5 m) carried by an unmanned ground vehicle using its two arms (two agents with total 14 joints). The robot needs to carry the bolts to the target position and the target orientation, and then, install bolts, as is shown in Fig. 3 a. The positions of two bolts are randomized in a fixed area (0.4 m × 0.24 m, the blue platform in Fig. 3 a) for each episode. Each arm is tasked to pick a fixed bolt throughout the whole process.

During the RL training process, when the distance between the bolts and the center point of gripper tips is less than a threshold (0.1 m in our study) and the orientation of the grippers are aligned (the center of the bolt, the center of tips, and the center of the palm are collinear), we assume the robot can pick up the bolt using simple inverse kinematics calculation and the task for RL is considered completed then. In addition, because the distance from the bolts to the bolt holes is small, we assume we can simply apply inverse kinematics to accurately install the bolt without considering the self-collisions anymore in such a short distance (self-collisions happen when agents approximate target positions).

To accelerate training, inverse kinematics is not employed and the agents just receive the rewards when they meet the conditions to trigger inverse kinematics. However, we will demonstrate the successful application of inverse kinematics of seamlessly working after RL control during the testing stage.

Task 1 can evaluate the performance of our MARL framework in (1) preventing self-collisions: the policies of agents are changing during the training, which poses a challenge for collision avoidance, and some initial positions of the bolts also result in difficulties in preventing self collisions; (2) working collaboratively for the same task to improve the efficiency.

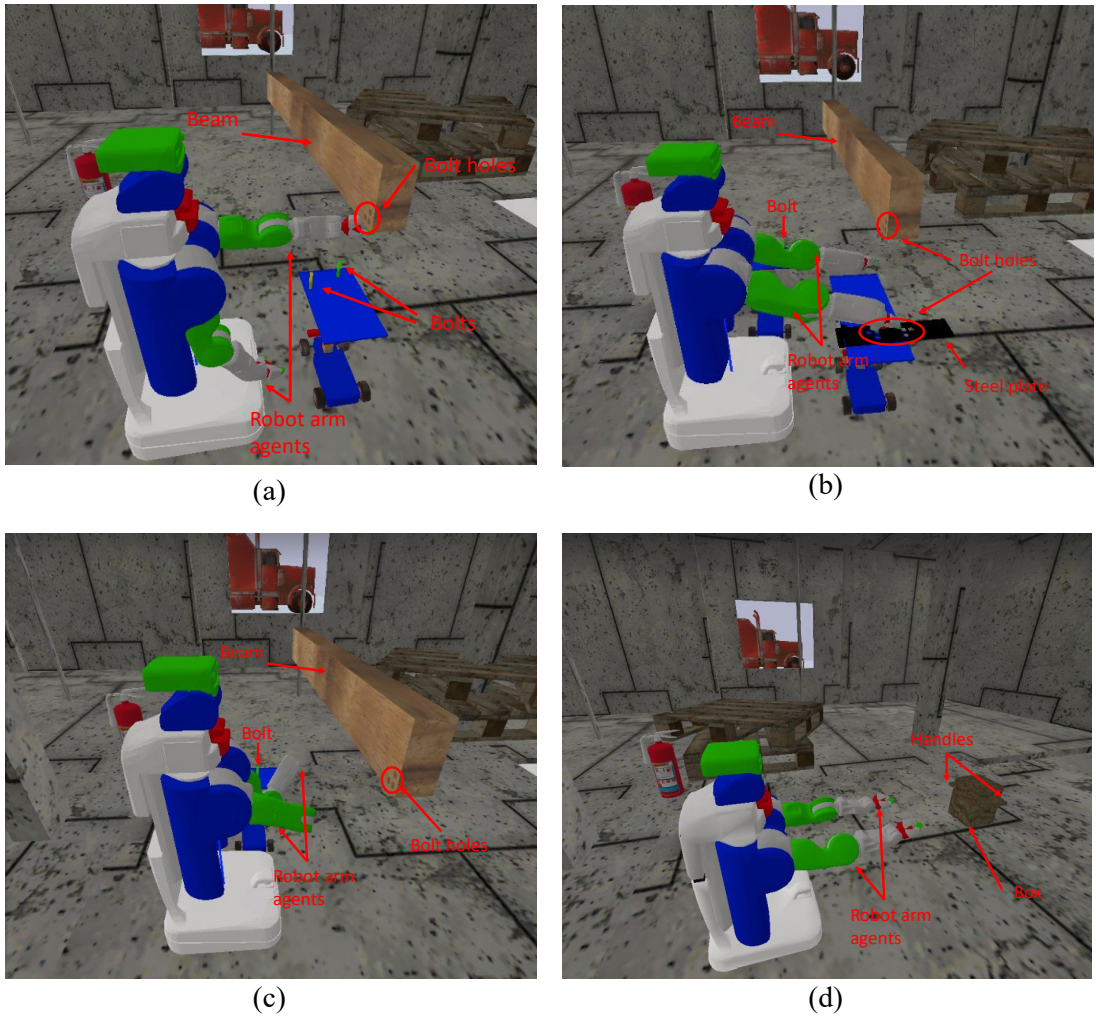

Fig. 3 Multi-agent pick and place task scenario.

*Task 2* Multi-agent beam installation task a (Fig. 3 b): The robot is asked to pick up a steel plate in which some holes are processed for fixation and use the bolt to fix the plate on the beam. The robot first picks up these two objects from their corresponding platforms and carries the plate and bolt to the target position. Then, the robot should align the plate with the beam and insert the bolt into the plate to fix the plate. Since it is difficult to pick up a plate on the surface with two fingers, we add a handle to the plate. The gripper thus can grasp the handle to carry and install the plate. The threshold distance for picking and installation of the RL part is the same as those in task 1. Inverse kinematics is applied after the RL part is completed. In addition, the RL should also rotate the plate to the desired orientation otherwise collisions might occur when rotating such a big plate near the beam using inverse kinematics. The initial position of bolts is randomized in a 0.1 m × 0.24 m area and that of the plate is randomized in a 0.05 m × 0.24 m area. In this task, due to the non-negligible volume of the steel plate, the agents should avoid collisions between the plate and other objects (e.g., bolts, platforms, and the beam) and work collaboratively to install the plate on the beam.

*Task 3* Multi-agent beam installation task b (Fig. 3 c): The robot needs to use its left arm to pick up the bolt on the left side and pass it to the right arm while the right arm needs to receive the bolt from its left end effector and then, install it on the beam. The threshold distance for picking and installation of the RL part is the same as those in task 1. The initial position of bolts is randomized in a 0.1 m × 0.2 m area. By collaboratively employing two arms in this task, the robot can work in a larger space compared with the case where only one arm is used. Since the policies of both agents are stochastic, two agents must work collaboratively in the dynamic environment so that they can efficiently pass the bolt from one hand to another hand.

*Task 4* Multi-agent box transportation task (Fig. 3 d): The robot needs to approximate the box with its four wheels (8 joints, 1 agent) and two arms, and then, move its arms to the handles of the box so that it can apply inverse kinematics to move upwards its two arms to carry the box. The grasp areas of the handles are defined by two tags added to the handles. The threshold distance for holding the handle is 0.1 m. The initial position of the box is randomized in a 1 m × 1 m area. Reinforcement learning is used to provide a generalized solution to approximate the box and put its end effector to the positions where we can easily apply inverse kinematics to lift the box. In this task, there are three agents and each of them has a high degree of freedom. Compared with the aforementioned tasks, the search space in this task is significantly larger since the robot can use its wheels to move anywhere. Applying such a complex environment can evaluate the learning efficiency of our MARL framework.

**4.2 Actions, observations, and rewards**

The actions of robot arms are controlled by the angle positions of each revolute joint as well as the orientation of the four wheels, and the movement of wheels is controlled by the rotation speed (each arm has 7 revolute joints and four wheels have 8 revolute joints). Observations should include all state information that is necessary to finish the task.

*Task 1:* each agent knows the actions of all agents, the position and rotation of grippers, the position of bolts, and the target positions. The reward function considers the distance from grippers to objects and from grippers to target positions as well as the collisions between the arms, and between the arms and objects, which can be written as:

$$r(t) = -\varphi_1 d_O(t) - \varphi_2 d_T(t) + \varphi_3 v_{og}(t) v_{gb}(t) - \varphi_4 c_s(t) - \varphi_5 c_o(t) - \varphi_6 h_g(t) + r_O(t) + r_T(t) \quad (6)$$

Where $\varphi_i$ denotes the weight coefficient for each term ($\varphi_1 = 0.2, \varphi_2 = 0.2, \varphi_3 = 0.1, \varphi_4 = 1.0, \varphi_5 = 1.0, \varphi_6 = 0.2$); $d_O$ and $d_T$ represent the distance from grippers to the object and the distance from the object to target positions, respectively; $v_{og}$ and $v_{gb}$ are vectors from the tip of the grippers to the object and the vector from the base of the grippers to the tip, respectively; $c_s$ and $c_o$ denote the self-collision and the collisions between robots and the objects, respectively; $r_O$ and $r_T$ are set to 1.0 when the robot picks up the object and places the object in to target position, respectively.

*Task 2:* the state information is similar to that in task 1, except that the agent should also observe the rotation of the plates as well as the positions of the holes on the plate. The reward function can be written as:

$$r(t) = -\varphi_1 d_O(t) - \varphi_2 d_T(t) + \varphi_3 v_{og}(t)v_{gb}(t) - \varphi_4 c_s(t) - \varphi_5 c_o(t) - \varphi_6 v_o(t) + r_O(t) + r_T(t) \quad (7)$$

Where $\varphi_i$ denotes the weight coefficient for each term ($\varphi_1 = 0.2, \varphi_2 = 0.2, \varphi_3 = 0.1, \varphi_4 = 1.0, \varphi_5 = 1.0, \varphi_6 = -0.1$); $d_O$ and $d_T$ represent the distance from the gripper to the objects (bolts and steel plates) and the distance from the objects to the target position, respectively; $v_o(t)$ denotes the difference between the current orientation of the plate and the target orientation. Other coefficients have the same representation as those in Eq. (6).

*Task 3:* the state information is similar to that in task 1, except that there is only one object in this scenario. The reward function considers the distance from grippers to objects and from grippers to target positions as well as the collisions between the arms and between the arms and objects, which can be written as:

$$r_l(t) = -\varphi_1 d_O(t) - \varphi_2 d_T(t) + \varphi_3 v_{og}(t)v_{gb}(t) - \varphi_4 c_s(t) - \varphi_5 c_o(t) + \varphi_6 v_{gl}(t)v_{gr}(t) + r_O(t) + r_T(t) \quad (8)$$

$$r_r(t) = -\varphi_1 d_O(t) - \varphi_2 d_T(t) + \varphi_6 v_{gl}(t)v_{gr}(t) - \varphi_4 c_s(t) - \varphi_5 c_o(t) + r_O(t) + r_T(t) \quad (9)$$

$r_r$ and $r_l$ denote the reward functions for the right-arm and left-arm agents, respectively. $\varphi_i$ denotes the weight coefficient for each term ($\varphi_1 = 0.2, \varphi_2 = 0.2, \varphi_3 = 0.1, \varphi_4 = 1.0, \varphi_5 = 1.0, \varphi_6 = 0.1$). In Eq. (8), we consider the distance between the bolt and the left end effector ($d_O$), and once the bolt is picked, the distance between the left effector to the right effector ($d_T$). In Eq. (9), $d_O$ is $d_T$ in Eq. (8), and $d_T$ represent the distance from the bolt to the target position. $v_{gl}$ and $v_{gr}$ are vectors represent the orientation of end effectors thus we can control the transfer of items from the left hand to the right hand. Once the right end effector picks up the bolt, the left arm successfully places the bolts.

*Task 4:* each agent knows the actions of all agents, the position and rotation of grippers, the position and rotation of the box, and the positions of the handles of the box. The reward function considers the distance from grippers to target positions as well as the collisions, which can be written as:

$$r_l(t) = r_r(t) = -\varphi_1 d_T(t) - \varphi_2 c_s(t) - \varphi_3 d_{o,o'}(t) + r_T(t) \quad (10)$$

$$r_w(t) = -\varphi_1 d_T(t) - \varphi_2 c_s(t) - \varphi_4 c_o(t) + r_T(t) \quad (11)$$

$r_w$ denotes the reward function of the wheel agent. $\varphi_i$ denotes the weight coefficient for each term ($\varphi_1 = 0.1, \varphi_2 = 0.5, \varphi_3 = 0.05, \varphi_4 = 1.0, \varphi_5 = 1.0$); $d_T$ represent the distance from grippers to target positions; $c_s$ and $c_o$ denote the self-collision and the collisions between the robot base and the objects, respectively; $d_{o,o'}$ reflects the movement of the box from its previous position; $r_T$ are set to 1.0 when the robot can lift the box using its arms.

**4.3 Configurations of training**

All four tasks are adopted for testing, and the configurations are shown in Table 2. The difference in configurations for tasks lies in the number of iterations and the maximum episode length.

Table 2 Training configurations.

|  | Train 1 | Train 2 | Train 3 | Train 4 |
|---|---|---|---|---|
| $N$ | 2 000 000 | 2 500 000 | 2 000 000 | 1 000 000 |
| $M$ | 2000 | 2000 | 2000 | 4000 |
| $n$ | 5 | 5 | 5 | 5 |
| $K$ | 4 | 4 | 4 | 4 |
| $lr$ | 0.0005 | 0.0005 | 0.0005 | 0.0005 |
| $\zeta$ | 0.2 | 0.2 | 0.2 | 0.2 |
| $\lambda_1$ | 0.5 | 0.5 | 0.5 | 0.5 |
| $\lambda_2$ | 0.01 | 0.01 | 0.01 | 0.01 |
| $\xi$ | 0.95 | 0.95 | 0.95 | 0.95 |
| $\gamma$ | 0.99 | 0.99 | 0.99 | 0.99 |
| Episode length | 20 | 30 | 25 | 50 |

## 5. Results and discussions

The learning curves are shown in Fig. 4. All learning curves show remarkable learning performance and convergence speed. Both agents in each task have learned the skill to finish the task, and when the left arm and right arm share the same reward function (e.g., Eq. 6), the gap between the learning performance of the two agents is not significant. Otherwise, the gap in learning curves between different agents depends on the configurations of the reward function and the construction environment.

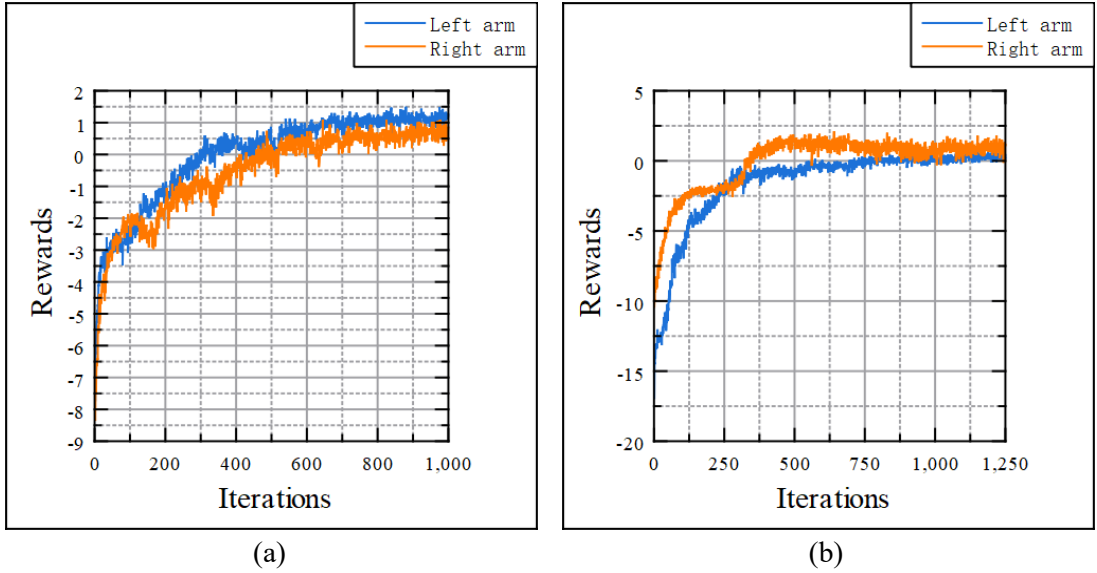

(a)          (b)

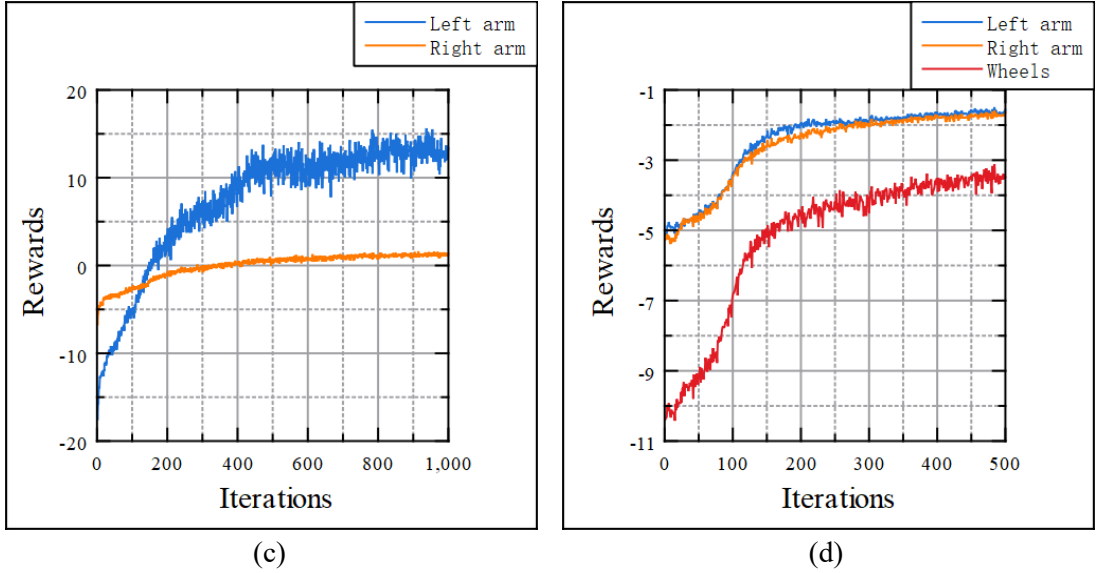

(c)                      (d)

Fig. 4 The learning curves of tasks.

Fig. 5 shows the learning results of task 1 with inverse kinematics used for installing the bolts. Both of its arms move toward the objects in a manner avoiding collisions. In Fig. 5, the initial positions of bolts pose challenges for the pick-up task since the right arm needs to grasp the bolt on the left side while the left arm needs to approximate the bolt on the right side. Therefore, although both arms try to approximate their bolts, they have to make sure their grippers will not conflict with each other, as shown in Fig. 5 (a)~(c). First, the left gripper picks up the bolt and rotates the gripper so that the right gripper can pick up another bolt. After that, the arm agents begin to move away from each other, as shown in Fig. 5 (c)~(e). When it comes to the installation phase, the agents need to place the object in adjacent positions, one arm hovers around the target position, and when one robot arm successfully places the object using inverse kinematics, another arm begins to install the object, as shown in Fig. 5 (e)~(f). With RL for collision-free path planning and inverse kinematics for installation, the robot can finish the task with desirable accuracy.

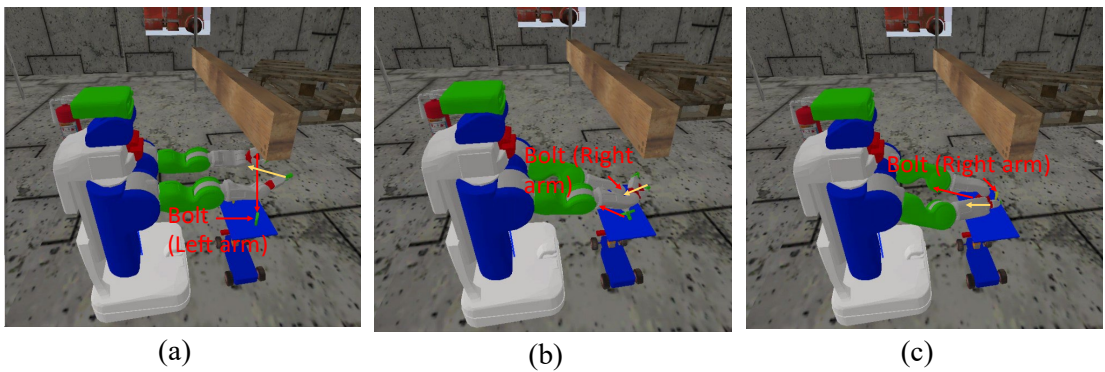

(a)                      (b)                      (c)

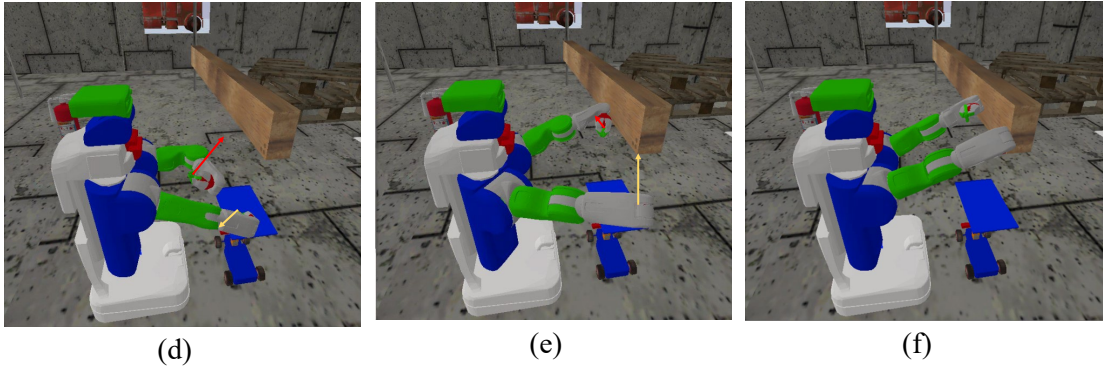

Fig. 5 Learning results of task 1.

Fig. 6 shows the learning results of task 2. The robot first moves its two arms to the objects and picks up them, as shown in Fig. 6 (a)~(b). Then, the right arm rotates its gripper while avoiding collisions between the platform and its steel plate (Fig. 6 c and d). After picking up the objects, the agents start to move toward the target positions. After that, inverse kinematics is applied to accurately align the plate and install the bolt, as shown in Fig. 6 (e)~(f).

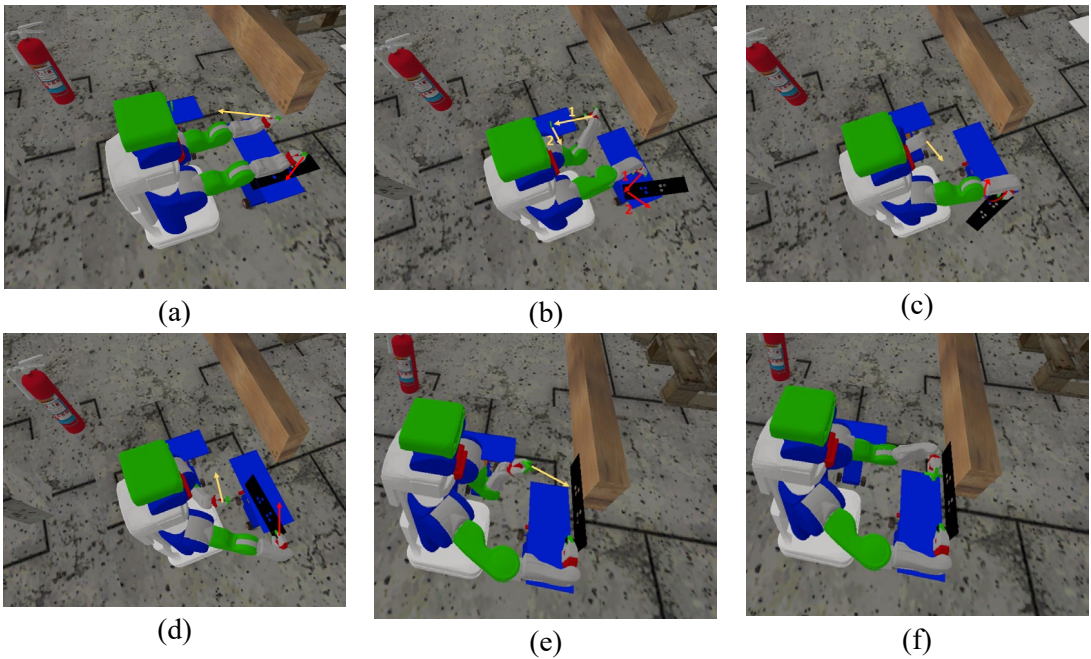

Fig. 6 Learning results of task 2.

Fig. 7 shows the learning results of task 3. The robot first moves its left arm towards the bolt, as shown in Fig. 7 (a)~(b). The right arm also moves towards the left arm to reduce the waiting time while avoiding self-collisions. After the left arm picks up the bolt, the robot transfers the bolt to its right arm, and then, each arm agent begins to move away from the other to avoid self collisions and the right arm carries the bolt to the target bolt hole to finish the installation, as shown in Fig. 7 (c)~(d).

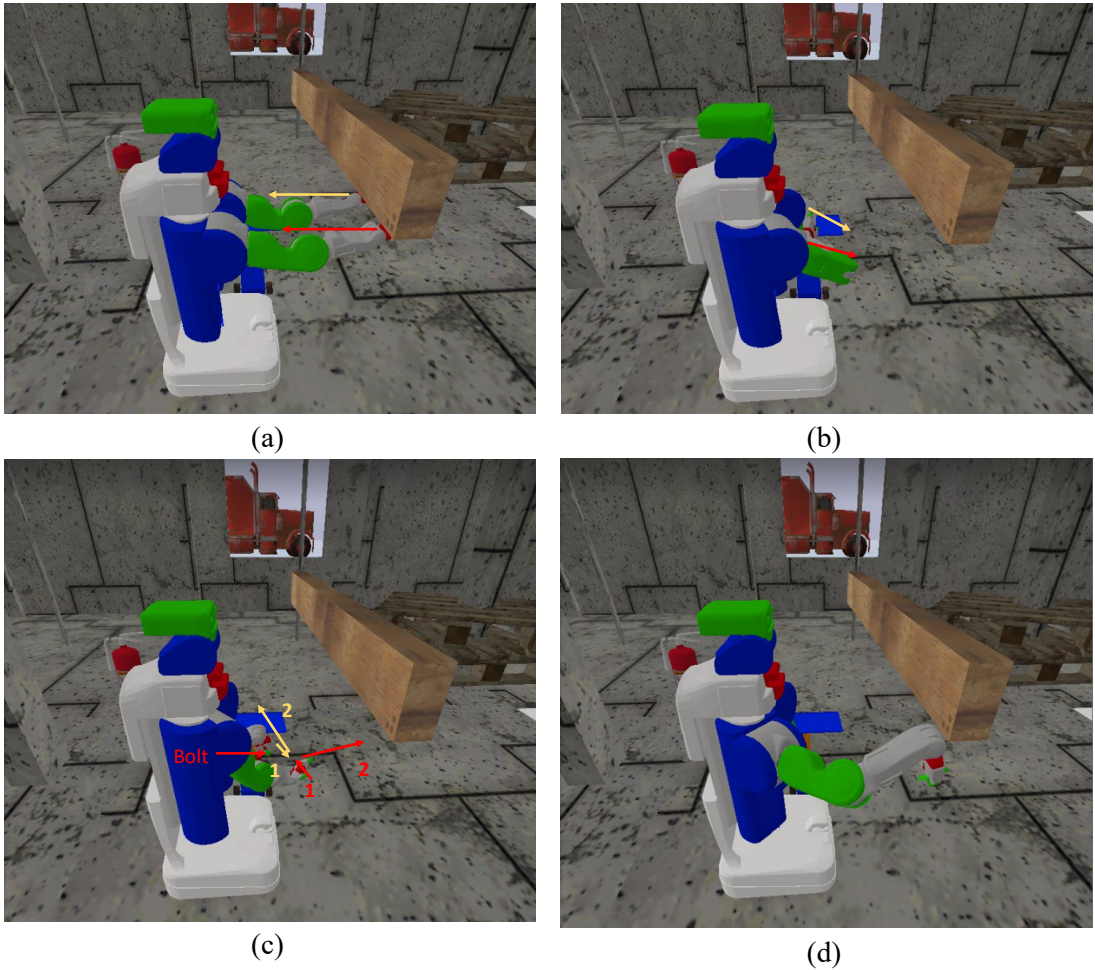

Fig. 7 Learning results of task 3.

In task 4, the robot first utilizes its four wheels to approximate the box with its two arms pointing to the handles, as shown in Fig. 8 (a)~(b). To prevent the box from being overturned by two arm agents and the wheel agent, the robot stops before the box and spread out its two arms, as shown in Fig. 8 (c). Then it moves its arms below the handles, as shown in Fig. 8 (d). Since the robot can approximate the box from any orientation and position, the arm agents require a strong generalization ability to move to the target positions.

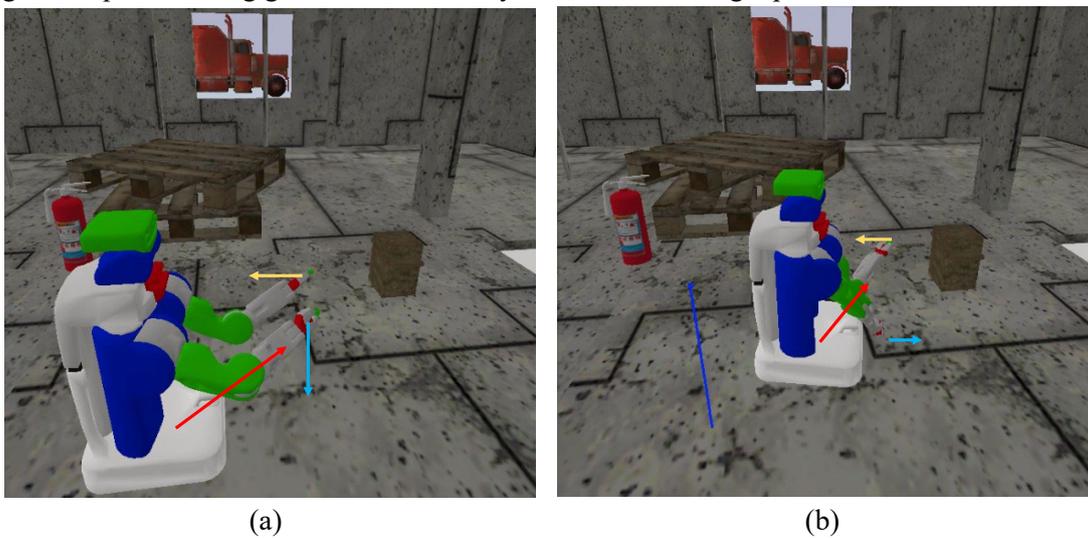

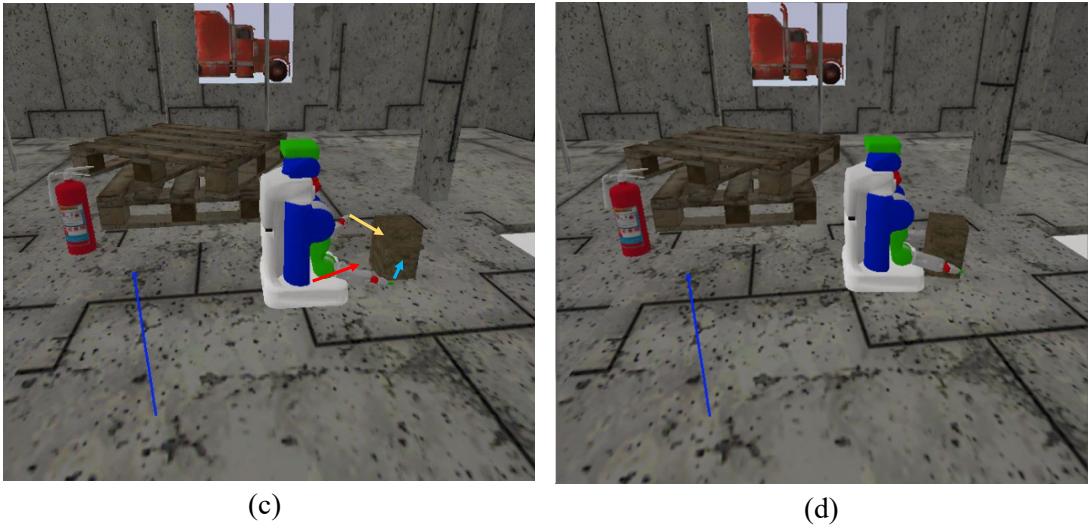

(c)                  (d)

Fig. 8 Learning results of task 4.

To test the learning results, we run the learned policies in four environments 100 times with randomized initial positions and get the average performance of intelligent agents. In tasks 1~3, the arm agent first needs to pick up the bolt and then, install or transfer the bolt, so we use two indexes to represent the success rate. In task 4, the robot only needs to hold the handles so one index is employed. In addition, we record the number of collisions per episode. In the collaborative installation tasks (tasks 1~3), collision avoidance is one of the key indexes of evaluation. All agents show great performance in task completion and collision avoidance (Table 3). When RL policies complete the approach task (i.e., move to a fixed position near the objects while avoiding collisions), inverse kinematics can always be applied to finish "the last 10 cm" (finish the installation or lifting precisely).

Table 3 The performance of RL policies.

| Task | Pick-up rate (%) | | Placement rate (%) | | Self-collisions per episode |
|---|---|---|---|---|---|
| | Left arm | Right arm | Left arm | Right arm | |
| 1 | 92 | 89 | 86 | 85 | 0.02 |
| 2 | 96 | 89 | 92 | 81 | 0.00 |
| 3 | 93 | 91 | 91 | 74 | 0.04 |
| | Success rate (%) | | | | |
| | Left arm | | Right arm | | |
| 4 | 84 | | 88 | | 0.00 |

## 6. Conclusions

In this paper, we present a novel MARL framework for collaborative or competitive construction robot control. Our framework utilizes MARL algorithms for trajectory planning and incorporates inverse kinematics for precise installation. By employing this framework, we enable the efficient completion of complex construction tasks that involve collaboration and competition among multiple agents. The main contributions of our work can be summarized as follows:

1. In this study, we first proposed a framework for multi-agent construction robot control by introducing a concise version of multi-agent PPO and successfully applying it to collaboration installation and material transportation scenarios. Our multi-agent PPO algorithm demonstrated excellent convergence speed and high learning performance in controlling multiple manipulators.
2. To facilitate future research in this field, we have designed a set of learning environments tailored specifically for construction robots within our framework. These environments encompass crucial components of a construction setting, including realistic robot models, diverse instruments, and lifelike construction scenes. We also explained the approach for extending the current construction instrument families in the environment. Moreover, our framework is highly adaptable and can seamlessly accommodate various MARL construction scenarios.
3. By leveraging the strengths of reinforcement learning and inverse kinematics, our framework enables robots to navigate complex multi-agent environments while avoiding collisions and achieving exceptional accuracy in installation tasks.
4. The versatility and adaptability of our framework were exemplified through the successful achievement of construction activities such as collaborative pick and place operations, bolt installation, and box lifting. The results showcased the remarkable learning performance achieved when controlling multiple agents with a high degree of freedom, enabling seamless collaboration in completing construction tasks.

Limitations exist in two aspects: one is that the adversarial scenario is relatively simple at the current stage; another is that the communication manner between agents can be improved. We will continue to work on this topic to improve these limitations in the future.

## Acknowledgments

Kangkang acknowledges the support of the CSC scholarship from the Chinese Scholarship Council.

## Data Availability Statement

The open-source project can be found on Github after publication.

## References:
[1] N. Melenbrink, J. Werfel andA. Menges, On-site autonomous construction robots: Towards unsupervised building, Automat Constr 119(2020) 103312.10.1016/j.autcon.2020.103312
[2] N. Melenbrink, K. Rinderspacher, A. Menges andJ. Werfel, Autonomous anchoring for robotic construction, Automat Constr 120(2020) 103391.10.1016/j.autcon.2020.103391
[3] S.K. Baduge, S. Thilakarathna, J.S. Perera, M. Arashpour, P. Sharafi, B. Teodosio, A. Shringi andP. Mendis, Artificial intelligence and smart vision for building and construction 4.0: Machine and deep learning methods and applications, Automat Constr 141(2022) 104440.10.1016/j.autcon.2022.104440
[4] H. Golizadeh, C.K.H. Hon, R. Drogemuller andM. Reza Hosseini, Digital engineering potential in addressing causes of construction accidents, Automat Constr 95(2018) 284-295.10.1016/j.autcon.2018.08.013
[5] X. Huang andJ. Hinze, Analysis of Construction Worker Fall Accidents, J Constr Eng M 129(2003) 262-271.10.1061/(ASCE)0733-9364(2003)129:3(262)
[6] X. Ma, C. Mao andG. Liu, Can robots replace human beings? —Assessment on the developmental potential of construction robot, Journal of Building Engineering 56(2022) 104727.10.1016/j.jobe.2022.104727

[7] C. Liang, X. Wang, V.R. Kamat andC.C. Menassa, Human‐Robot Collaboration in Construction: Classification and Research Trends, J Constr Eng M 10(2021) 3121006
[8] E. Gambao, C. Balaguer andF. Gebhart, Robot assembly system for computer-integrated construction, Automat Constr 9(2000) 479-487.10.1016/S0926-5805(00)00059-5
[9] K. Jung, B. Chu andD. Hong, Robot-based construction automation: An application to steel beam assembly (Part II), Automat Constr 32(2013) 62-79.10.1016/j.autcon.2012.12.011
[10] B. Chu, K. Jung, M. Lim andD. Hong, Robot-based construction automation: An application to steel beam assembly (Part I), Automat Constr 32(2013) 46-61.10.1016/j.autcon.2012.12.016
[11] T. Sasaki andK. Kawashima, Remote control of backhoe at construction site with a pneumatic robot system, Automat Constr 17(2008) 907-914.10.1016/j.autcon.2008.02.004
[12] R. Saltaren, R. Aracil andO. Reinoso, Analysis of a Climbing Parallel Robot for Construction Applications, Comput-Aided Civ Inf 19(2004) 436-445.10.1111/j.1467-8667.2004.00368.x
[13] M. Vega-Heredia, R.E. Mohan, T.Y. Wen, J.S. Aisyah, A. Vengadesh, S. Ghanta andS. Vinu, Design and modelling of a modular window cleaning robot, Automat Constr 103(2019) 268-278.10.1016/j.autcon.2019.01.025
[14] S. Lee andT.M. Adams, Spatial Model for Path Planning of Multiple Mobile Construction Robots, Comput-Aided Civ Inf 19(2004) 231-245.10.1111/j.1467-8667.2004.00351.x
[15] D. Lee andM. Kim, Autonomous construction hoist system based on deep reinforcement learning in high-rise building construction, Automat Constr 128(2021) 103737.10.1016/j.autcon.2021.103737
[16] R.S. Sutton andA.G. Barto, Reinforcement Learning: An Introduction, The MIT Press2017.
[17] V. Mnih, K. Kavukcuoglu, D. Silver, A. Graves, I. Antonoglou andD.W.M. Riedmiller, Playing Atari with Deep Reinforcement Learning, arXiv2013)
[18] J. Schulman, F. Wolski, P. Dhariwal, A. Radford andO. Klimov, Proximal Policy Optimization Algorithms, arXiv2017)
[19] T.P. Lillicrap, J.J. Hunt, A. Pritzel, N. Heess, T. Erez, Y. Tassa, D. Silver andD. Wierstra, Continuous control with deep reinforcement learning, arXiv2015)
[20] B. Belousov, B. Wibranek, J. Schneider, T. Schneider, G. Chalvatzaki, J. Peters andO. Tessmann, Robotic architectural assembly with tactile skills: Simulation and optimization, Automat Constr 133(2022) 104006.10.1016/j.autcon.2021.104006
[21] A.A. Apolinarska, M. Pacher, H. Li, N. Cote, R. Pastrana, F. Gramazio andM. Kohler, Robotic assembly of timber joints using reinforcement learning, Automat Constr 125(2021) 103569.10.1016/j.autcon.2021.103569
[22] S.R.B. Dos Santos, S.N. Givigi andC.L. Nascimento, Autonomous construction of structures in a dynamic environment using Reinforcement Learning, in: IEEE International Systems Conference, IEEE, 2013,452-459.
[23] H. Li andH. He, Multiagent Trust Region Policy Optimization, IEEE Trans Neural Netw Learn Syst PP(2023).10.1109/TNNLS.2023.3265358
[24] R. Lowe, Y. Wu, A. Tamar, J. Harb, P. Abbeel andI. Mordatch, Multi-Agent Actor-Critic for Mixed Cooperative-Competitive Environments, arXiv:1706.022752020)
[25] Y. Wang, H. Liu, W. Zheng, Y. Xia, Y. Li, P. Chen, K. Guo andH. Xie, Multi-Objective Workflow Scheduling With Deep-Q-Network-Based Multi-Agent Reinforcement Learning, Ieee Access 7(2019) 39974-39982.10.1109/ACCESS.2019.2902846
[26] T. Rashid, M. Samvelyan, C.S. de Witt, G. Farquhar, J. Foerster andS. Whiteson, Monotonic Value Function Factorisation for Deep Multi-Agent Reinforcement Learning, J Mach Learn Res2020) 1-51
[27] H. Choi, C. Han, K. Lee andS. Lee, Development of hybrid robot for construction works with pneumatic actuator, Automat Constr 14(2005) 452-459.10.1016/j.autcon.2004.09.008
[28] S. Kang andE. Miranda, Planning and visualization for automated robotic crane erection processes in construction, Automat Constr 15(2006) 398-414.10.1016/j.autcon.2005.06.008
[29] V.R. Konda andJ.N. Tsitsiklis, Actor-Critic Algorithms, in: Advances in Neural Information Processing Systems, 1999.
[30] F.A. Azad, S.A. Rad andM. Arashpour, Back-stepping control of delta parallel robots with smart dynamic model selection for construction applications, Automat Constr 137(2022) 104211.10.1016/j.autcon.2022.104211
[31] C. Liang, V.R. Kamat andC.C. Menassa, Teaching robots to perform quasi-repetitive construction tasks through human demonstration, Automat Constr 120(2020) 103370.10.1016/j.autcon.2020.103370

[32] L. Huang, Z. Zhu andZ. Zou, To imitate or not to imitate: Boosting reinforcement learning-based construction robotic control for long-horizon tasks using virtual demonstrations, Automat Constr 146(2023) 104691.10.1016/j.autcon.2022.104691

[33] V. Asghari, A.J. Biglari andS. Hsu, Multiagent Reinforcement Learning for Project-Level Intervention Planning under Multiple Uncertainties, J Manage Eng 2(2023) 4022075

[34] C.P. Andriotis andK.G. Papakonstantinou, Managing engineering systems with large state and action spaces through deep reinforcement learning, Reliab Eng Syst Safe 191(2019) 106483.10.1016/j.ress.2019.04.036

[35] L. Yu, Z. Xu, T. Zhang, X. Guan andD. Yue, Energy-efficient personalized thermal comfort control in office buildings based on multi-agent deep reinforcement learning, Build Environ 223(2022) 109458.10.1016/j.buildenv.2022.109458

[36] J. Vazquez-Canteli, T. Detjeen, G. Henze, J. Kämpf andZ. Nagy, Multi-agent reinforcement learning for adaptive demand response in smart cities, Journal of Physics: Conference Series 1343(2019) 12058.10.1088/1742-6596/1343/1/012058

[37] K. Miyazaki, N. Matsunaga andK. Murata, Formation path learning for cooperative transportation of multiple robots using MADDPG, in: ICROS, 2021,1619-1623.

[38] J. Liu, P. Liu, L. Feng, W. Wu, D. Li andY.F. Chen, Automated clash resolution for reinforcement steel design in concrete frames via Q-learning and Building Information Modeling, Automat Constr 112(2020) 103062.10.1016/j.autcon.2019.103062

[39] R. Diankov, Automated Construction of Robotic Manipulation Programs, in: Carnegie Mellon University, Robotics Institute, Pittsburgh, 2010.

[40] S. Huang andS. Ontañón, A Closer Look at Invalid Action Masking in Policy Gradient Algorithms, in: Cornell University Library, arXiv.org, Ithaca, 2022.

[41] J. Schulman, P. Moritz, S. Levine, M. Jordan andP. Abbeel, HIGH-DIMENSIONAL CONTINUOUS CONTROL USING GENERALIZED ADVANTAGE ESTIMATION, arXiv:1506.024382018)

[42] E. Coumans, Bullet Physics Simulation, in: SIGGRAPH '15, New York, NY, USA, 2015.